\documentclass[a4paper]{article}
\pdfoutput=1
\usepackage[utf8]{inputenc}
\usepackage{geometry}
\usepackage{bibentry}
\usepackage{bbm}
\usepackage{amsmath}
\usepackage[cm]{fullpage}
\usepackage{xspace}
\usepackage{amsmath,amssymb,amsfonts,amsthm}
\usepackage{hyperref}
\usepackage[round]{natbib}
\theoremstyle{definition}
\newtheorem{theorem}{Theorem}
\newtheorem{lemma}[theorem]{Lemma}
\newtheorem{claim}[theorem]{Claim}
\newtheorem{example}[theorem]{Example}

\allowdisplaybreaks

\DeclareMathOperator{\Tr}{tr}
\newcommand{\ie}[0]{\emph{i.e.},\xspace}

\title{A Note on Bounding Regret of the \\ C$^2$UCB Contextual Combinatorial Bandit}
\author{Bastian Oetomo \and Malinga Perera \and Renata Borovica-Gajic \and Benjamin I. P. Rubinstein\footnote{School of Computing and Information Systems, University of Melbourne, Parkville, VIC 3010, Australia.  \texttt{$\{$boetomo,wpperera$\}$@student.unimelb.edu.au}, \texttt{$\{$rborovica,brubinstein$\}$@unimelb.edu.au}}}

\date{\today}

\begin{document}

\maketitle

\begin{abstract}
We revisit the proof by \citet{qin2014contextual} of bounded regret of the C$^2$UCB contextual combinatorial bandit. We demonstrate an error in the proof of volumetric expansion of the moment matrix, used in upper bounding a function of context vector norms. We prove a relaxed inequality that yields the originally-stated regret bound.
\end{abstract}

\section{Introduction}\label{sec:intro}
In deriving a regret bound on the C$^2$UCB contextual combinatorial bandit, \citet{qin2014contextual} use the following equality within the proof of their Lemma~4.2. 
\begin{claim}
    \label{claim:incorrect}
    Let $k, m, n$ be natural numbers, $\mathbf{V}$ be a $d \times d$ real and positive definite matrix, and $S_t \subseteq [m]$ with $|S_t|\leq k\leq m$ for $t\in[n]$. Let $\mathbf{x}_t(i)\in\mathbb{R}^d$ be vectors for $t\in[n], i\in[m]$, and define $\mathbf{V}_n = \mathbf{V} + \sum_{t=1}^n\sum_{i\in S_t}\mathbf{x}_t(i)\mathbf{x}_t(i)^T$. Then
    $\det(\mathbf{V}_n) = \det(\mathbf{V})\prod_{t=1}^n\left(1+\sum_{i\in S_t}\|\mathbf{x}_t(i)\|^2_{\mathbf{V}_{t-1}^{-1}}\right)$,
    where we define $\|\mathbf{a}\|_{\mathbf{M}}=\sqrt{\mathbf{a}^T\mathbf{M}\mathbf{a}}$.
\end{claim}

We present a counterexample to Claim~\ref{claim:incorrect} in Section~\ref{sec:counter}, and then in Section~\ref{sec:proof} prove the relaxation given by,

\begin{lemma}\label{lemma:amain} Under the same conditions as Claim~\ref{claim:incorrect},
    $\det(\mathbf{V}_n) \geq \det(\mathbf{V})\prod_{t=1}^n\left(1+\sum_{i\in S_t}\|\mathbf{x}_t(i)\|^2_{\mathbf{V}_{t-1}^{-1}}\right)$.
\end{lemma}

In the setting of C$^2$UCB, $[n], [m]$ correspond to rounds and arms, $S_t$ the (super arm of) played arms in round $t$, $x_t(i)$ the context vector for arm $i$ at round $t$, and $V_t$ the covariance matrix from the played contexts added to $V$ (taken to be a scaled identity, for achieving ridge regression reward estimates). 
We detail in Section~\ref{sec:implications} how Lemma~\ref{lemma:amain} can be used within the remainder of the proof of \citep[Lemma 4.2]{qin2014contextual}, ultimately yielding the C$^2$UCB regret bound originally claimed. The regret analysis of C$^2$UCB is based on previous analysis of contextual bandits \citep{auer2002using,dani2008stochastic,chu2011contextual}. We 
 demonstrate that the bound in Lemma~\ref{lemma:amain} is sharp, by describing conditions for equality. 

\paragraph{Notation.} We denote by $\lambda_i(\mathbf{A})$ the eigenvalues of the $n \times n$ matrix $\mathbf{A}$, where, without loss of generality, $\lambda_1(\mathbf{A}) \leq \lambda_2(\mathbf{A}) \leq \dots \leq \lambda_n(\mathbf{A})$.
We likewise order $S_t = \{ s_{(1,t)}, s_{(2,t)}, \dots, s_{(|S_t|,t)}\},$ where $s_{(1,t)} < s_{(2,t)} < \dots < s_{(|S_t|,t)}$. 

\paragraph{Generalised Matrix Determinant Lemma.}
We make use of the identity: 
Let $\mathbf{A}$ be an invertible $n \times n$ matrix, and $\mathbf{B}, \mathbf{C}$ be $n \times m$ matrices, then
$\det(\mathbf{A}+\mathbf{B}\mathbf{C}^T) = \det(\mathbf{I}_m+\mathbf{C}^T\mathbf{A}^{-1}\mathbf{B})\det(\mathbf{A})$.
\ifdefined\longpap
\emph{Proof.}
\begin{align*}
    \det(\mathbf{A}+\mathbf{B}\mathbf{C}^T) &= \det(\mathbf{A}+\mathbf{B}\mathbf{C}^T) \det(\mathbf{A}^{-1})\det(\mathbf{A})\\
    &= \det(\mathbf{I}_n+\mathbf{B}\mathbf{C}^T\mathbf{A}^{-1})\det(\mathbf{A})\\
    &= \det(\mathbf{I}_n+\mathbf{B}(\mathbf{C}^T\mathbf{A}^{-1}))\det(\mathbf{A})\\
    &= \det(\mathbf{I}_m + \mathbf{C}^T\mathbf{A}^{-1}\mathbf{B})\det(\mathbf{A})
\end{align*}
where the last line is due to the Sylvester's determinant identity, with $\mathbf{B}$ being an $n \times m$ matrix and $\mathbf{C}^T\mathbf{A}^{-1}$ being an $m \times n$ matrix.
\fi

\ifdefined\longpap
\paragraph{Eigenvalues of sum of a matrix and the identity,}
For a real symmetric $n \times n$ matrix $\mathbf{A}$ with eigenvalues $\lambda_1(\mathbf{A}), \dots, \lambda_n(\mathbf{A})$, the eigenvalues of $(\mathbf{I}_n+\mathbf{A})$ is $(1+\lambda_1(\mathbf{A})), \dots, (1+\lambda_n(\mathbf{A}))$.\\
\emph{Proof.}\\
Any real symmetric matrix $\mathbf{A}$ is always diagonalisable, which can be written as $\mathbf{A}=\mathbf{P}\mathbf{D}\mathbf{P}^{-1}$, where $\mathbf{D}$ is a diagonal matrix consisting the eigenvalues of $\mathbf{A}$ and $\mathbf{P}$ consisting of the corresponding eigenvectors. Therefore,
\begin{align*}
    \mathbf{I}_n+\mathbf{A} &= \mathbf{P}\mathbf{I}_n\mathbf{P}^{-1} + \mathbf{P}\mathbf{D}\mathbf{P}^{-1}\\
    &= \mathbf{P}(\mathbf{I}_n+\mathbf{D})\mathbf{P}^{-1}.
\end{align*}
The last line tells us that $\mathbf{I}_n+\mathbf{A}$ is diagonalisable with its eigenvalues listed diagonally in the matrix $(\mathbf{I}_n+\mathbf{D})$. Therefore, $\lambda_i(\mathbf{I}_n+\mathbf{A}) = \lambda_i(\mathbf{A})+1$.
\fi

\section{A Counterexample}\label{sec:counter}

Claim~\ref{claim:incorrect} derives from the assertion
within the proof of~\citep[Lemma 4.2]{qin2014contextual} that,
$$\det(\mathbf{V}_{n-1})\det\left(\mathbf{I}+\sum_{i\in S_n}(\mathbf{V}_{n-1}^{-1/2}\mathbf{x}_n(i))(\mathbf{V}_{n-1}^{-1/2}\mathbf{x}_n(i))^T\right) = \det(\mathbf{V}_{n-1})\det\left(\mathbf{I}+\sum_{i\in S_n}\|\mathbf{x}_n(i)\|^2_{\mathbf{V}_{n-1}^{-1}}\right)\;.$$
This appears to conflate outer and inner products, after basis transformation by $\mathbf{V}_{n-1}^{-1/2}$. The following counterexample to Claim~\ref{claim:incorrect} establishes that
indeed it does not hold in general.

\begin{example}
Consider $n = 1$, the $2\times 2$ matrix $\mathbf{V} = 1.2 \mathbf{I}_2$, $S_t = \{1,2,3\}$ and let $
    \mathbf{x}_1(1)=
    \begin{bmatrix}
        0.3\\
        0.7
    \end{bmatrix}, \quad
    \mathbf{x}_1(2)=
    \begin{bmatrix}
        0.6\\
        0.1
    \end{bmatrix}, \quad
    \mathbf{x}_1(3) = 
    \begin{bmatrix}
        0.1\\
        0.5
    \end{bmatrix}.$
It follows that $\mathbf{V}_1 =
\begin{bmatrix}
    1.66 &0.32\\
    0.32 &1.95
\end{bmatrix}$.
Then we have
\begin{align*}
    &\det(\mathbf{V})\prod_{t=1}^n\left(1+\sum_{i\in S_t}\|\mathbf{x}_t(i)\|^2_{\mathbf{V}_{t-1}^{-1}}\right) \\
    =& \det(\mathbf{V})\left(1+\sum_{i=1}^3\mathbf{x}_1(i)^T\mathbf{V}^{-1}\mathbf{x}_1(i)\right)\\
    =& \det\left(1.2\mathbf{I}_2\right)\left(1+\mathbf{x}_1(1)^T\left(\frac{1}{1.2}\mathbf{I}\right)\mathbf{x}_1(1)+\mathbf{x}_1(2)^T\left(\frac{1}{1.2}\mathbf{I}\right)\mathbf{x}_1(2)+\mathbf{x}_1(3)^T\left(\frac{1}{1.2}\mathbf{I}\right)\mathbf{x}_1(3)\right)\\
    =& 1.2^2 \left(1+\frac{1}{1.2}(0.3^2+0.7^2)+\frac{1}{1.2}(0.6^2+0.1^2) + \frac{1}{1.2}(0.1^2+0.5^2)\right)\\
    =& 2.892 \; \neq \; 3.1346 \; =\; 1.66\times 1.95-0.32\times 0.32 \; = \; \det(\mathbf{V}_1)\;.
\end{align*}
\end{example}

\section{Proof of Lemma~\ref{lemma:amain}}\label{sec:proof}
Let $\mathbf{X}_{n} = \begin{bmatrix}
    \mathbf{x}_{n}(s_{(1,n)}) &\dots &\mathbf{x}_{n}(s_{(|S_{n}|,n)})
\end{bmatrix}$. Then,
\begin{align*}
    \det(\mathbf{V}_n)
    &= \det\left(\mathbf{V} + \sum_{t=1}^n\sum_{i\in S_t}\mathbf{x}_t(i)\mathbf{x}_t(i)^T\right)\\
    &= \det\left(\mathbf{V} + \sum_{t=1}^{n-1}\sum_{i\in S_t}\mathbf{x}_t(i)\mathbf{x}_t(i)^T +\sum_{i\in S_n} \mathbf{x}_n(i)\mathbf{x}_n(i)^T\right)\\
    &= \det\left(\mathbf{V}_{n-1} + \mathbf{X}_{n}\mathbf{X}_{n}^T\right)\\
    &= \det(\mathbf{V}_{n-1})\det\left(\mathbf{I}_{|S_{n}|}+\mathbf{X}_{n}^T\mathbf{V}_{n-1}^{-1}\mathbf{X}_{n}\right)\\
    &= \det(\mathbf{V}_{n-1})\left[ \prod_{i=1}^{|S_{n}|}\lambda_i\left(\mathbf{I}_{|S_{n}|}+\mathbf{X}_{n}^T\mathbf{V}_{n-1}^{-1}\mathbf{X}_{n}\right)\right]\\
    &= \det(\mathbf{V}_{n-1})\left[ \prod_{i=1}^{|S_{n}|}\left(1+\lambda_i\left(\mathbf{X}_{n}^T\mathbf{V}_{n-1}^{-1}\mathbf{X}_{n}\right)\right)\right] \;,
\end{align*}
where the fourth and final equalities follow from the Generalised Matrix Determinant Lemma and the fact that adding the identity to a square matrix increases eigenvalues by one. 
Now, the final line's product can be expanded as
\begin{align}
1 + \sum_{i=1}^{|S_{n}|} \lambda_i(\mathbf{X}_{n}^T\mathbf{V}_{n-1}^{-1}\mathbf{X}_{n}) + \sum_{1 \leq i_1 < i_2 \leq |S_{n}|} \lambda_{i_1}(\mathbf{X}_{n}^T\mathbf{V}_{n-1}^{-1}\mathbf{X}_{n})\lambda_{i_2}(\mathbf{X}_{n}^T\mathbf{V}_{n-1}^{-1}\mathbf{X}_{n})+\dots + \prod_{i=1}^{|S_{n}|}\lambda_i(\mathbf{X}_{n}^T\mathbf{V}_{n-1}^{-1}\mathbf{X}_{n})\;. \label{eq:prodexpanse}
\end{align}
Since $\mathbf{V}$ is positive definite and $\mathbf{x}_t(i)\mathbf{x}_t(i)^T$ is positive semi-definite (with one eigenvalue being $\mathbf{x}_t(i)^T \mathbf{x}_t(i)$ and the remainder all zero) for all $t$ and $i$, we have that $\mathbf{V}_{n-1} = \mathbf{V} + \sum_{t=1}^{n-1}\sum_{i\in S_t}\mathbf{x}_t(i)\mathbf{x}_t(i)^T$ is positive definite. 
Therefore, we conclude that $\mathbf{V}_{n-1}^{-1}$ is also positive definite, hence it has a symmetric square root matrix $\mathbf{V}_{n-1}^{-1/2}$. It also follows that $\mathbf{X}_{n}^T\mathbf{V}_{n-1}^{-1}\mathbf{X}_{n}$ is positive semi-definite. 
\ifdefined\longpap
This is because  for any $|S_n| \times 1$ vector $\mathbf{a}$, where $\mathbf{a}\neq \mathbf{0}$,
\begin{align*}
    \mathbf{a}^T(\mathbf{X}_n^T\mathbf{V}_{n-1}^{-1}\mathbf{X}_n)\mathbf{a} &= \mathbf{a}^T\mathbf{X}_n^T\mathbf{V}_{n-1}^{-1/2}\mathbf{V}_{n-1}^{-1/2}\mathbf{X}_n\mathbf{a} \\
    &= \mathbf{a}^T\mathbf{X}_n^T\mathbf{V}_{n-1}^{-T/2}\mathbf{V}_{n-1}^{-1/2}\mathbf{X}_n\mathbf{a}\\
    &= (\mathbf{V}_{n-1}^{-1/2}\mathbf{X}_n\mathbf{a})^T(\mathbf{V}_{n-1}^{-1/2}\mathbf{X}_n\mathbf{a})\\
    &= \|\mathbf{V}_{n-1}^{-1/2}\mathbf{X}_n\mathbf{a}\|^2\\
    &\geq 0,
\end{align*}
which shows that all the eigenvalues of $\mathbf{X}_n^T\mathbf{V}_{n-1}^{-1}\mathbf{X}_n$ are non-negative.
\fi
Therefore, the terms starting from the third term in the expansion \eqref{eq:prodexpanse} are all non-negative because they are products of the eigenvalues of $\mathbf{X}_{n}^T\mathbf{V}_{n-1}^{-1}\mathbf{X}_{n}$. Thus we have,
\begin{align*}
    \det(\mathbf{V}_{n})
    &= \det(\mathbf{V}_{n-1})\left[ \prod_{i=1}^{|S_{n}|}\left(1+\lambda_i\left(\mathbf{X}_{n}^T\mathbf{V}_{n-1}^{-1}\mathbf{X}_{n}\right)\right)\right]\\
    &\geq \det(\mathbf{V}_{n-1})\left(1 + \sum_{i=1}^{|S_{n}|} \lambda_i(\mathbf{X}_{n}^T\mathbf{V}_{n-1}^{-1}\mathbf{X}_{n})\right)\\
    &= \det(\mathbf{V}_{n-1})\left( 1 + \Tr(\mathbf{X}_{n}^T\mathbf{V}_{n-1}^{-1}\mathbf{X}_{n})\right) \\
     &= \det(\mathbf{V}_{n-1})\left( 1 + \sum_{i\in S_{n}} \mathbf{x}_{n}(i)\mathbf{V}_{n-1}^{-1}\mathbf{x}_{n}(i)\right)\\
     &= \det(\mathbf{V}_{n-1})\left( 1 + \sum_{i\in S_{n}} \|\mathbf{x}_{n}(i)\|^2_{\mathbf{V}_{n-1}^{-1}}\right)\; ,
\end{align*}
where the third equality follows from expanding out the argument to the trace as 
\begin{align*}
    \mathbf{X}_n^T\mathbf{V}_{n-1}^{-1}\mathbf{X}_n
    &= \begin{bmatrix}
        \mathbf{x}_n(s_{(1,n)})^T \mathbf{V}_{n-1}^{-1} \mathbf{x}_n(s_{(1,n)}) &\dots & \mathbf{x}_n(s_{(1,n)})^T \mathbf{V}_{n-1}^{-1}\mathbf{x}_n(s_{(|S_n|,n)})\\
        \vdots &\ddots &\vdots\\
        \mathbf{x}_n(s_{(|S_n|,n)})^T \mathbf{V}_{n-1}^{-1} \mathbf{x}_n(s_{(1,n)}) &\dots &\mathbf{x}_n(s_{(|S_n|,n)})^T \mathbf{V}_{n-1}^{-1}\mathbf{x}_n(s_{(|S_n|,n)})
    \end{bmatrix}\;.
\end{align*}
Applying our recurrence relation on $\mathbf{V}_t$ for $1\leq t \leq n$, we can telescope to arrive at
\ifdefined\longpap
\begin{align*}
    \det(\mathbf{V}_{n})
    &\geq \det(\mathbf{V}_{n-1})\left( 1 + \sum_{i\in S_{n}} \|\mathbf{x}_{n}(i)\|^2_{\mathbf{V}_{n-1}^{-1}}\right)\\
    &\geq \det(\mathbf{V}_{n-2})\left( 1 + \sum_{i\in S_{n-1}} \|\mathbf{x}_{n-1}(i)\|^2_{\mathbf{V}_{n-2}^{-1}}\right)\left( 1 + \sum_{i\in S_{n}} \|\mathbf{x}_{n}(i)\|^2_{\mathbf{V}_{n-1}^{-1}}\right)\\
    &\geq \dots\\
    & \geq \det(\mathbf{V})\prod_{t=1}^n\left(1+\sum_{i\in S_t}\|\mathbf{x}_t(i)\|^2_{\mathbf{V}_{t-1}^{-1}}\right),
\end{align*}
which proves 
\fi
the result.

\section{Implication of Lemma~\ref{lemma:amain}}\label{sec:implications}
By rearranging the inequality, we know that
$$\prod_{t=1}^n\left(1+\sum_{i\in S_t}\|\mathbf{x}_t(i)\|^2_{\mathbf{V}_{t-1}^{-1}}\right) \leq \dfrac{\det(\mathbf{V}_n)}{\det(\mathbf{V})}\;,$$ provided that $\det(\mathbf{V}) > 0$, which is guaranteed for our positive definite $\mathbf{V}$. 
The next steps of \citep[Lemma 4.2]{qin2014contextual}'s proof follow the original pattern\footnote{Here as in the original proof, we leverage assumptions:  $\lambda_1(V)\geq k$ and the context vectors are of bounded norm $\|\mathbf{x}_t(i)\|_2\leq 1$.} now with the second inequality in what follows (due to our Lemma~\ref{lemma:amain} and monotonicity), rather than the original equality:
\begin{align*}
    \sum_{t=1}^n\sum_{i\in S_t}\|\mathbf{x}_t(i)\|^2_{\mathbf{V}_{t-1}^{-1}}
    &\leq 2\sum_{t=1}^n\log\left(1+\sum_{i\in S_t} \|\mathbf{x}_t(i)\|^2_{\mathbf{V}_{t-1}^{-1}}\right)
    = 2\log \left[ \prod_{t=1}^n \left(1+\sum_{i\in S_t}\|\mathbf{x}_t(i)\|^2_{\mathbf{V}_{t-1}^{-1}}\right)\right] \\
    &\leq 2\log\left( \dfrac{\det(\mathbf{V}_n)}{\det(\mathbf{V})}\right) 
    = 2 \log(\det(\mathbf{V}_n)) - 2\log(\det(\mathbf{V})),
\end{align*}
which yields the regret bound as presented by \citet{qin2014contextual}, without further modification to the proof of their Lemma 4.2.

\section{Discussion}\label{sec:discussion}
The proof of Lemma~\ref{lemma:amain} offers intuition as to when the inequality holds with equality. Namely, it is true when the matrix $\mathbf{X}_{t}^T\mathbf{V}_{t-1}^{-1}\mathbf{X}_{t}$ has at most one non-zero eigenvalue \ie be either a rank-1 or rank-0 matrix for all $1 \leq t \leq n$. This is because the terms that we dropped in calculating the determinant of $\mathbf{I}_{|S_{t}|}+\mathbf{X}_{t}^T\mathbf{V}_{t-1}^{-1}\mathbf{X}_{t}$ are then identically $0$. This agrees with the result of the non-generalised matrix determinant lemma.

This occurs when intra-round, played context vectors are co-linear to each other: 
if the context vector of arm $i$ can be written as $\mathbf{x}_t(i)=a_{it} \mathbf{u}_t$, then we can write $\mathbf{X}_t = \mathbf{u}_t\mathbf{a}_t^T$, where $\mathbf{a}_t$ is a column vector with $a_{it}$ as its components. The matrix we are interested in becomes $\mathbf{X}_t^T\mathbf{V}_{t-1}^{-1}\mathbf{X}_t = (\mathbf{u}_t\mathbf{a}_t^T)^T\mathbf{V}_{t-1}^{-1}(\mathbf{u}_t\mathbf{a}_t^T) = \|\mathbf{u}_t\|^2_{\mathbf{V}_{t-1}^{-1}}\mathbf{a}_t\mathbf{a}_t^T$, which is a rank-1 matrix. Thus, it also follows that the trace of this matrix is $\|\mathbf{u}_t\|^2_{\mathbf{V}_{t-1}^{-1}}\|\mathbf{a}_t\|^2$. One interesting thing to notice here is that the context vectors need not to be co-linear across rounds.

A special case of the co-linearity scenario is the non-combinatorial bandit. In this scenario, $|S_t| = 1$ for all $t$. This means that given a particular round $t$, there is only 
one context vector available. In particular, $\det\left(\mathbf{I}_{|S_{t}|}+\mathbf{X}_{t}^T\mathbf{V}_{t-1}^{-1}\mathbf{X}_{t}\right) = \det\left(\mathbf{I}_{1}+\mathbf{x}_{t}^T\mathbf{V}_{t-1}^{-1}\mathbf{x}_{t}\right) = 1+\mathbf{x}_{t}^T\mathbf{V}_{t-1}^{-1}\mathbf{x}_{t}$, which is the bound that we had for calculating $\det\left(\mathbf{I}_{|S_{t}|}+\mathbf{X}_{t}^T\mathbf{V}_{t-1}^{-1}\mathbf{X}_{t}\right)$, were $|S_t|=1$.

\bibliography{biblio}

\end{document}